\documentclass[journal]{IEEEtran}
\usepackage{hyperref}
\usepackage{url}
\usepackage{subfig}
\usepackage{amsmath}
\usepackage{amsfonts}
\usepackage{amssymb}
\usepackage{multirow}
\usepackage{graphicx}
\usepackage{bm}
\usepackage{cleveref}
\usepackage{epstopdf}
\usepackage{algorithm}
\usepackage[noend]{algpseudocode}
\usepackage{pifont}
\usepackage{psfrag}

\ifCLASSOPTIONcompsoc
  \usepackage[nocompress]{cite}
\else
  \usepackage{cite}
\fi

\ifCLASSINFOpdf
\else
\fi

\DeclareMathOperator{\argmax}{argmax}
\DeclareMathOperator{\argmin}{argmin}

\DeclareMathOperator{\E}{\mathbb{E}}

\DeclareMathOperator{\D}{\mathcal{D}}

\renewcommand{\H}{\mathcal{H}}

\DeclareMathOperator{\R}{\mathbb{R}}
\DeclareMathOperator{\X}{\mathcal{X}}
\DeclareMathOperator{\Y}{\mathcal{Y}}
\DeclareMathOperator{\Z}{\mathcal{Z}}

\let\oldReturn\Return
\renewcommand{\Return}{\State\oldReturn}

\begin{document}
\title{Detection of Adversarial Training Examples in Poisoning Attacks through Anomaly Detection}

\author{Andrea~Paudice,
        Luis~Mu\~{n}oz-Gonz\'alez,
        Andr\'as~Gy\"orgy,
        and~Emil~C. Lupu
\IEEEcompsocitemizethanks{
\IEEEcompsocthanksitem A. Paudice, L. Mu\~{n}oz-Gonz\'alez, and  E.C. Lupu are with the Department
of Computing, Imperial College London, United Kingdom,
180 Queen's Gate, SW7 2AZ, London. A. Gy\"orgy is with the Department of Electrical and Electronic Engineering, Imperial College London, United Kingdom, SW7 2BT.\protect\\
E-mail: \{a.paudice15, l.munoz, a.gyorgy, e.c.lupu\}@imperial.ac.uk}}


\IEEEtitleabstractindextext{%
\begin{abstract}
Machine learning has become an important component for many systems and applications including computer vision, spam filtering, malware and network intrusion detection, among others. Despite the capabilities of machine learning algorithms to extract valuable information from data and produce accurate predictions, it has been shown that these algorithms are vulnerable to attacks. Data poisoning is one of the most relevant security threats against machine learning systems, where attackers can subvert the learning process by injecting malicious samples in the training data. Recent work in adversarial machine learning has shown that the so-called optimal attack strategies can successfully poison linear classifiers, degrading the performance of the system dramatically after compromising a small fraction of the training dataset. In this paper we propose a defence mechanism to mitigate the effect of these optimal poisoning attacks based on outlier detection. We show empirically that the adversarial examples generated by these attack strategies are quite different from genuine points, as no detectability constrains are considered to craft the attack. Hence, they can be detected with an appropriate pre-filtering of the training dataset.
\end{abstract}

}

\maketitle

\IEEEdisplaynontitleabstractindextext

\IEEEpeerreviewmaketitle

\section{Introduction}
In the big data era an increasing number of services rely on data-driven approaches that leverage the information available from many sources, including devices, sensors and people. Machine learning algorithms allow to extract valuable information from this overwhelming amount of data and have been successfully applied in many application domains. This trend is set to continue as machine learning is increasingly adopted to support important economical assets. Cyber-security applications are also increasingly relying on machine learning systems, and machine learning is one of the main components of state-of-the-art systems for detecting malware, spam, or network intrusions \cite{stolfo2004detecting, shin2015recognizing, Newsome2006}. 
However, machine learning systems are vulnerable themselves  \cite{laskov2009framework, skillicorn2009adversarial, laskov2010machine, barreno2010security, tygar2011adversarial, huang2011adversarial, biggio2014security}, and can be targeted by attackers. In particular an attacker can exploit their vulnerabilities to avoid detection or to degrade the performance of the system by injecting malicious data. Far from a mere hypothesis, such attacks have already been reported \emph{in the wild} against autonomous \emph{bots},\footnote{https://www.wired.com/2017/02/keep-ai-turning-racist-monster/} anti-virus engines,\footnote{https://www.virusbulletin.com/uploads/pdf/conference\_slides/2013/\\BatchelderJia-VB2013.pdf} spam filters, as well as to create fake news\footnote{https://www.wired.com/2017/02/internet-made-fake-news-thing-made-nothing/} or profiles in social networks, among others. Among others, the emerging field of \emph{adversarial machine learning} aims at understanding the security properties of machine learning by both analyzing the mechanisms that can allow attackers to compromise the algorithms and proposing defence techniques to mitigate them \cite{huang2011adversarial}. 

We can broadly distinguish between two different kind of attacks: \emph{evasion} and \emph{poisoning} attacks \cite{barreno2010security}. In the first case, the attacker aims to produce an intentional error in the machine learning system to perpetrate malicious activities or to cause misclassification at run-time, i.e., when the system is already deployed. For example, \cite{hosseini2017} shows how to evade Google's \emph{Perspective} API\footnote{https://www.perspectiveapi.com/} to detect toxic comments by replacing words that are likely to be identified as offensive with slightly altered versions of those words. In a \textit{poisoning} attack, the attacker is assumed capable of partially modifying the training data used by the learning algorithm, producing a \emph{bad} model and causing a degradation of the system's performance, which may facilitate, among others, subsequent system evasion. For example, in the context of spam filtering, attackers can craft spam emails that contain words likely to appear in good emails. When the spam filter is retrained on new examples including these malicious emails, the learned filter will increase significantly its false positive rate \cite{nelson2008exploiting}. 

Poisoning attacks are considered one of the most significant threats for systems that rely upon collecting data in the wild \cite{joseph2013}. For example, many services use the users' feedback on their decisions to update or refine the learning algorithm. An attacker can thus provide malicious feedback crafted to poison the system and compromise its performance. The integrity of data collection and labelling are often not guaranteed in many applications. For example, malware samples are often collected from compromised machines with known vulnerabilities - like \emph{honeypots} - or from online services like VirusTotal,\footnote{https://virustotal.com} where labelling errors are frequent. Previous work \cite{li2016data, alfeld2016data, huang_icml15, mei2015using, biggio12-icml} has demonstrated that popular machine learning algorithms can be successfully poisoned with well-crafted adversarial examples. The proposed attack strategies have often focused on a \emph{worst-case analysis}, where the poisoning examples are designed to maximise the damage on the learning algorithm in a targeted or an indiscriminate way. However, these attack strategies only consider the learning algorithm, and the restrictions about the poisoning points that can be generated are loose and arbitrary. Usually, the attack points are only required to be within the feasible domain of the feature space. 

Defence techniques against poisoning attacks have been less explored in the research community. In \cite{nelson2008exploiting}, the proposed defence strategy relies on measuring the impact of each sample on the performance of the algorithm. Although it is effective to detect some poisoning attacks, testing the impact of each sample in the training dataset is often computationally very intensive, as it requires to re-train and evaluate the performance of the algorithm for every sample in the set. Even if we consider only a subset of the samples for re-training, this can be impractical for many machine learning applications. 
In \cite{feng2014robust}, the authors propose a defence strategy that eliminates outliers by solving an optimization problem that aims to identify suspicious samples. However, the algorithm requires to estimate the percentage of data points that are expected to be outliers (which should be provided as a parameter), which is far from a trivial task. If the estimate does not corresponds to the reality the performance of the algorithm is significantly affected, especially if the system is not under attack. Furthermore, outliers are assumed to be independent and identically distributed which is usually not the case of smart attack strategies, where the adversarial points are generated together to maximize the power of the attack.

In this paper we propose a different approach to detect adversarial examples in poisoning attacks by considering the whole pipeline of typical machine learning systems. The so-called ``optimal'' attacks proposed in \cite{huang_icml15, mei2015using, biggio12-icml} only consider the learning algorithm, overlooking previous data pre-processing steps (the attacks are called optimal since they select poisoning examples to maximize the damage to the learner). We show that the adversarial examples generated by these methods can be effectively detected and filtered with appropriate outlier detection techniques. The \emph{curse of dimensionality} \cite{keogh2011curse} implies that it is hard to estimate the underlying probability distribution of the data in high-dimensional spaces (especially if the number of data points is limited). Thus, applying outlier detection under these settings is far from a trivial task. Notwithstanding these limitations, our experimental results show that it is possible to mitigate the effect of the attacks with outlier detection, even in the cases where the data is scarce compared to the number of features. Our experimental evaluation also shows that although more constrained attack strategies, such as label flipping attacks, are less effective than ``optimal'' attacks, when considering the learning algorithm in isolation, the adversarial examples generated are much harder to detect. Thus, label flipping attacks are more harmful when considering the whole machine learning pipeline. This suggests the necessity of designing more realistic attack strategies to provide more accurate worst-case analysis by analyzing the entire machine learning system, not just the learning algorithm, in line with the sophistication that can be displayed by smart attackers.

The main contributions of the paper are the following:
\begin{itemize}
\item We propose an effective and computationally efficient countermeasure for poisoning attacks against linear classifiers based on data pre-filtering with outlier detection. This defence strategy is agnostic w.r.t. the actual learning algorithm and provides a useful methodology to enforce the security of practical machine learning systems against poisoning attacks.
\item An experimental evaluation on real datasets, showing the effectiveness of our proposed defence methodology, including examples where the number of features is high compared to the number of training points. The experimental results show that the effect of the ``optimal'' poisoning strategies described in \cite{huang_icml15, mei2015using, biggio12-icml} is strongly mitigated by our countermeasure. 
\item We show in our experiments that simpler constrained attacks, such as label flipping, are much harder to detect. We also provide an empirical comparison of the effectiveness of ``optimal'' and label flipping attacks when considering the machine learning in isolation and when we apply our defence technique. To the best of our knowledge, this is the first systematic study of the benefits of outlier detection to mitigate the effect of data poisoning against machine learning algorithms. 
\end{itemize}

The rest of the paper is organized as follows: In Section~2 we describe the related work. In Section~3 we introduce the ``optimal'' poisoning attack strategy formulation in the context of linear models for binary classification. In Section~4 we describe our defence technique to detect adversarial training examples via outlier detection. In Section~5 we show our experimental results on synthetic and real datasets. Finally, in Section~6 we conclude the paper and discuss further research avenues.
\section{Related work}
Poisoning attacks have been mainly explored in the context of binary classification. The first practical poisoning attacks have been proposed in \cite{nelson2008exploiting,kloft2012security} for applications in spam filtering and anomaly detection. However, such attacks do not easily generalize to different learning algorithms. 
More systematic approaches have been proposed in \cite{biggio12-icml,huang_icml15,mei2015using,Munoz-Gonzalez:2017:TPD:3128572.3140451} where optimal attack strategies are defined as a bi-level optimization problem where the attacker aims at maximizing some objective function whilst the defender learns the parameters of the model by minimizing some cost function evaluated on the (tainted) training dataset. In particular, \cite{biggio12-icml} first described this attack strategy proposing a formulation to compromise support vector machines (SVMs). Following a similar approach, \cite{huang_icml15} showed the vulnerabilities of popular regularized learning algorithms and regularization-based embedded feature selection methods, such as \emph{ridge regression}, \emph{Lasso}, and \emph{elastic net}, against data poisoning. \cite{mei2015using} have considered a unified framework to poison convex learning algorithms with \emph{Tikhonov regularizers}. 
Finally, an alternative algorithm to solve the bi-level optimization problem based on back-gradient optimization is proposed in \cite{Munoz-Gonzalez:2017:TPD:3128572.3140451}. This algorithm computes efficiently the gradients required to solve the corresponding bi-level optimization problem through automatic differentiation. Compared to previous approaches, this technique does not require to replace the inner optimization problem with its first order optimality condition, whose solutions are only available for some learning algorithms in closed form. Then, back-gradient optimization allows to perform poisoning attacks against a broader range of algorithms, including neural networks and deep learning systems, extending the formulation of the optimal attack to multi-class classification problems. 
Other poisoning attacks have been devised against different supervised machine learning algorithms, such as \emph{collaborative filtering}, \emph{autoregressive models} for time-series forecasting, or linear regression \cite{li2016data,alfeld2016data,liu2011false,DBLP:journals/tsg/EsnaolaPPK16}. These approaches also consider the presence of a hypothesis-testing-based detector that can identify suspicious examples. Thus, the proposed poisoning strategies model this scenario, constraining the attack points to evade the anomaly detector. However, there is no empirical evaluation to support the validity and the performance of their approaches. 

On the defender's side, \cite{nelson2008exploiting} propose to measure the impact of each training example on the classifier's performance to detect poisoning points. Examples that affect negatively the performance are then discarded. Although this technique has been proven to be effective against some types of poisoning attacks, it is computationally very expensive and it can suffer from overfitting when the dataset is small compared to the number of features. In the context of generative models, \cite{feng2014robust} propose a two-step defence strategy for logistic regression classifiers: First outlier detection is applied, and then, the algorithm is trained solving an optimization problem based on the sorted correlation between the classifier and the labels. The main shortcoming is that the defender is assumed to know in advance the fraction of poisoning examples, which is not available in practical applications, and the performance of the algorithm is sensitive to this value. 
Recently, in \cite{pmlr-v70-koh17a} \emph{influence functions} from robust statistics have been proposed both as countermeasure to label flipping attacks and to generate poisoning examples. The idea is to identify the impact that each training example has on the classifier loss without retraining the model. In this way, it is possible to efficiently identify more dangerous examples that need a careful investigation from the user and to identify which data point to modify in order to maximally compromise the classifier's performance. A different approach to the problem has been taken in \cite{DBLP:journals/corr/SteinhardtKL17} where a data-dependent upper bound on the performance of the learner under data poisoning is approximated with an online algorithm. The authors assume that the learner performs some data sanitization before training, and  derive an optimal attack as bi-product of their upper bound estimation procedure. Note however that the provided upper bound is only on the expected surrogate risk rather than on the classification performance and may be quite loose.

We propose a new technique to detect adversarial training examples in linear classifiers through outlier detection, regardless of the training algorithm. As shown in Section~5, our approach is effective to mitigate the optimal attack strategies proposed in \cite{biggio12-icml,huang_icml15,mei2015using}. In contrast to \cite{nelson2008exploiting}, our defence technique is computationally more efficient and can be applied to large datasets. Furthermore, the algorithm does not require to know the fraction of poisoning points in advance, as it is the case in \cite{feng2014robust}. As we demonstrate in Section~5, it is also efficient in cases where the number of training points is of the same order as the number of features. However, more constrained attack strategies, like label flipping, cannot be effectively mitigated with our technique. We see our work as a complement to \cite{DBLP:journals/corr/SteinhardtKL17}, as we consider more effective outlier detection strategies against attackers which blindly optimize the learner's error. 
\section{Optimal Poisoning Attacks}
In a poisoning attack the adversary injects examples in the learner's training dataset to influence the behavior of the learning algorithm according to some goal defined by the attacker. Typically adversarial examples are designed to maximize the error of the learning algorithm, although more subtle objectives can also be considered. In line with most of the existing literature in poisoning attacks, we only consider here binary linear classification problems. Thus, in this section we describe the general framework to design optimal attack strategies as a bi-level optimization problem \cite{biggio12-icml,huang_icml15, mei2015using}. Then, we show a particular case for Lasso-like classifiers, which are widely used in many applications. Different levels of knowledge for the attacker can be considered, although we restrict our formulation to perfect knowledge attacks, where the attacker knows everything about the target system: both the learning algorithm (including the feature selection process) and the training data. Although unrealistic for practical scenarios, the assumption of perfect knowledge provides \emph{worst-case} evaluations of the robustness of machine learning systems. This can help to produce a quantitative estimate of the risk and of the maximum degradation of the system performance under the presence of a smart attacker, e.g. for applications that require assurance on the performance of the system. Furthermore, the evaluation of these attack strategies under the assumption of perfect knowledge can be applied as a testing benchmark to design more robust machine learning systems.

\subsection{Problem formulation}
Formally, a binary classification task is defined by a tuple $(\Z, \D, \H, \l)$: $\Z = \X \times \Y$ is the sample space, where $\X$ represents the domain and $\Y = \{-1, +1\}$ the labels; $\D$ is a probability distribution over $\Z$, $\H \subseteq \{f: \X \rightarrow \Y\}$ is a set of classifiers, and $\ell: \H \times \Z \rightarrow \R_+$ is a loss function. Given a training set $S_{tr}$\footnote{$S_{tr}$ is a multi-set, so repeated instances are allowed.} of $n$ i.i.d. samples from $\D$, the goal of a learning algorithm is to output a classifier $h_S : \X \rightarrow \Y$ such that $L_{\D}(h) = \E_{z \sim \D}[\ell(h_S, z)]$ is small. Many popular learning algorithms achieve this goal by minimizing some empirical estimate $O_L$ of $L_{\D}$, evaluated on the training set. Note that $O_L$ is typically the regularized empirical average of a surrogate loss $\ell'$ of $\ell$. In a poisoning attack, the attacker's goal is to find a set of $p$ poisoning points $S_p = \{ \textbf{z}_i \}_{i=1}^p = \{{\bf x}_{i}, y_{i} \}_{i=1}^p$ that maximizes some objective function $O_A$ when they are added to the training set, so that the new training set becomes $\hat{S}_{tr} = S_{tr} \cup S_p$, where $S_{tr}$ represents the untainted training set. Similarly to \cite{biggio12-icml,huang_icml15, mei2015using}, if we assume that the classifier is parametrized by ${\bf w}$, we can formulate the attacker's objective as the following bi-level optimization problem:
\begin{equation}
\begin{aligned}
S_p^* \in \underset{S_p \subseteq \Z}{\argmax}\quad & O_A=O_L(S_{val}, {\bf w}^*)\\
\text{s.t.}\quad & {\bf w}^* \in \underset{{\bf w}}{\argmin} \;  O_L(\hat{S}_{tr}, {\bf w}) \\
\end{aligned}
\label{eq3}
\end{equation} where the attacker's objective $O_A$ is evaluated in a separate validation set $S_{val}$. In previous studies \cite{biggio12-icml,huang_icml15, mei2015using}, the labels of the poisoning points, $y_{p_i}$, are assumed to be selected arbitrarily by the attacker and then fixed throughout the optimization process, where we only have to find the values $X_p = \{ \textbf{x}_i \}_{i=1}^p$ that maximize the objective $O_A$. 

Solving Problem (\ref{eq3}) is, in general, NP-Hard \cite{bard2013practical} and, even if the loss function used by the classifier is convex, the optimization problem is non-convex. However, it is possible to apply a gradient ascent strategy to get a (possibly) local maximum of the optimization problem in (\ref{eq3}). As in \cite{biggio12-icml,huang_icml15, mei2015using}, provided that the loss function $O_L$ and the learning algorithm are continuously differentiable w.r.t. the parameters ${\bf w}$ and the poisoning points $X_p$, we can replace the inner optimization problem in (\ref{eq3}) by the corresponding first order optimality condition (i.e., the gradient of $O_L$ is zero): 
\begin{equation}
\begin{aligned}
S_p^* \in \underset{S_p \in \Z}{\argmax}\quad & O_A(S_{val}, {\bf w}^*) \\
\text{s.t.}\quad & \frac{\partial O_L}{\partial {\bf w}^*}(\hat{S}_{tr}, {\bf w}^*) = {\bf 0} \\
\end{aligned}
\label{eq4}
\end{equation} Then, for the gradient ascent strategy, we can compute the following update rule:
\begin{equation}
X_p^{(t+1)} = \Pi_{\X} \left( X_p^{(t)} + \eta^{(t)}\frac{\partial O_A}{\partial X_p} \right)
\label{eq5}
\end{equation} where $\Pi_{\X}({\bf x})$ is a projection operator to project ${\bf x}$ onto the feasible domain $\X$, to handle bounded feature values 
This projection operator only ensures that the values of the poisoning points are valid points, but it does not include any further restriction to include detectability constraints. As we show in our experimental evaluation in Section~5, the lack of control on the values that the poisoning points can take results in poisoning points that are outliers that can be detected and pre-filtered.

The derivative of the attacker's objective function $O_A$ w.r.t. the poisoning points is usually not explicit: $O_A$ is evaluated on a separate validation set $S_{val}$, independent of the poisoning points. However, there is an implicit relation between $O_A$ and $X_p$ through the parameters learned by the classifier ${\bf w}^*$, i.e. by changing the value of the poisoning points we also change the value of ${\bf w}^*$, and so, the value of $O_A$. Thus, we can write the corresponding derivative in (\ref{eq5}) as:
\begin{equation}
\frac{\partial O_A}{\partial X_p} = \frac{\partial {\bf w}^*}{\partial X_p} \frac{\partial O_A}{\partial {\bf w}^*}
\label{eq6}
\end{equation} As suggested in \cite{mei2015using} we can apply the implicit function theorem on the KKT conditions in the inner optimization problem in (\ref{eq4}), so that:
\begin{equation}
\frac{\partial {\bf w}^*}{\partial X_p} = - \left( \frac{\partial^2 O_L}{\partial {\bf w} \partial {\bf w}^T} \right)^{-1} \frac{\partial^2 O_L}{\partial X_p \partial {\bf w}^T}
\label{eq7}
\end{equation} The application of the implicit function theorem in (\ref{eq7}) requires the expression of the KKT conditions to be differentiable w.r.t. both ${\bf w}$ and $X_p$. Moreover, it also requires the Hessian $\partial^2 O_L / (\partial {\bf w} \partial {\bf w}^T)$ to be a full rank matrix. Although these conditions may appear restrictive, this attack strategy can be successfully applied to well-known linear classifiers such as the perceptron, logistic regression, or linear SVM \cite{biggio12-icml,huang_icml15, mei2015using}.

\subsection{Linear regression for classification: poisoning lasso}
Similarly to \cite{huang_icml15}, in this section we describe the formulation of the optimal poisoning attack strategy for an $\ell_1$ regularized linear classifier. We use this classifier for our experimental evaluation in Sect.~5. As shown in \cite{Munoz-Gonzalez:2017:TPD:3128572.3140451}, we can expect a similar degradation of the performance of different linear classifiers against optimal poisoning attack. Then, the use of a different classifier should not impact the validity of our results. Assuming that the labels are $\Y = \{-1, 1\}$, the hypothesis predicted by the linear classifier for a sample ${\bf x} \in \R^d$ is given by $h({\bf x}) = \mbox{sign}({\bf w}^\top {\bf x} + b)$, where ${\bf w} \in \R^d$ and $b \in \R$ are the parameters of the classifier, and $d$ is the number of features. The simplest way to learn the parameters of this classifier is minimizing the Mean Squared Error (MSE) over the training set. Thus, in our poisoning scenario the loss function for the learning algorithm can be expressed as:
\begin{equation}
O_L(\hat{S}_{tr}, ({\bf w}, b)) = \frac{1}{2 n_{tr}} \sum_{i=1}^{n_{tr}} ({\bf w}^\top {\bf x}_i + b - y_i)^2 + \lambda \|{\bf w}\|_1
\label{eq8}
\end{equation} where $n_{tr}$ is the number of training samples and $\lambda \in \R^+$ is the regularization parameter. Because of the $\ell_1$ penalty term added to the cost function, the learned classifier is also known as lasso for classification \cite{tibshirani1996regression}. Note that by setting $\lambda = 0$ we recover the standard perceptron algorithm. Lasso-like classifiers are very popular in practical applications where the number of features is larger or comparable with the number of training points. In these scenarios standard algorithms are prone to \emph{overfitting}, i.e., the performance at training time is significantly better than the performance at test time. The $\ell_1$ regularization term helps to mitigate overfitting, reducing the complexity of the learned classifier by zeroing some of the components of ${\bf w}$. Therefore, classifiers with lasso are also referred as a particular type of embedded feature selection algorithms. 

Applying the gradient strategy previously described for an entire set of poisoning points can be computationally expensive. As in \cite{huang_icml15}, we use a strategy where we compute the gradient for one poisoning point at a time. Following the optimization problem defined in (\ref{eq4}), we can compute the derivative of the attacker's objective w.r.t. a single poisoning point ${\bf x}_p$ as:
\begin{equation}
\frac{\partial O_A}{\partial {\bf x}_p} = \frac{1}{n_{val}} \sum_{j=1}^{n_{val}} ({\bf w}^\top {\bf x}_j + b - y_j) \left( {\bf x}_j \frac{\partial {\bf w}}{\partial {\bf x}_p} + \frac{\partial {\bf w}}{\partial b} \right)
\label{eq9}
\end{equation} where $n_{val}$ is the number of validation sets used by the attacker. To solve (\ref{eq9}) we need to compute the implicit derivatives as in (\ref{eq7}), so that:
\begin{equation}
\begin{bmatrix}
\frac{\partial \bf{w}}{\partial \textbf{x}_p}	\\[0.3em]
\frac{\partial b}{\partial \textbf{x}_p} 		\\[0.3em]
\end{bmatrix}
=
-\frac{1}{n}
\begin{bmatrix}
\bm{\Sigma}		& \bm{\mu}			\\[0.3em]
\bm{\mu}^\top	&  1				\\[0.3em]
\end{bmatrix}^{-1}
\begin{bmatrix}
{\bf M}					\\[0.3em]
{\bf w}^\top					\\[0.3em]
\end{bmatrix}
\label{eq10}
\end{equation} where $\bm{\Sigma} =  \sum_{i = 1}^{n_{tr}}  {\bf x}_i {\bf x}_i^\top$, $\bm{\mu} = \sum_{i = 1}^{n_{tr}} {\bf x}_i$ is the sample mean of the training data and ${\bf M} = {\bf x}_p {\bf w}^\top + ({\bf x}_p^\top {\bf w} + b - 1){\bf I}_d$ with ${\bf I}_d$ being the $d \times d$ identity matrix.

Algorithm \ref{alg1} describes the whole procedure for the optimal poisoning attack strategy against Lasso-like classifiers. Although the algorithm is based on the attack strategy proposed in \cite{huang_icml15}, we have modified some of the steps, as we describe below. The algorithm starts initializing the $q$ poisoning examples that the attacker wants to craft (line 3). For this, several alternatives can be considered. Similarly to \cite{huang_icml15} we assume that $y_{p_j}$, the labels of the poisoning points, are arbitrarily set by the attacker. We train the classifier with the training set $S_{tr}$. Then, we select samples with labels opposite to $y_{p_j}$ so that the loss for these samples is maximized when flipping their labels to $y_{p_j}$. At each iteration, the algorithm updates the points individually (lines 8-11). In contrast to \cite{huang_icml15}, we retrain the classifier every time we compute the derivative for a poisoning point. This allows us to compute the gradient accurately, whereas the strategy proposed in \cite{huang_icml15} only produces an approximate gradient. Finally, we have also included a non-constant learning rate $\eta$ which is computed at each iteration applying the \emph{Golden Section} (GS) method described in \cite{gentle2009computational}. This allows to reduce the number of iterations and to enhance the stability of the gradient ascent algorithm. 

\begin{algorithm}
\begin{algorithmic}[1]
\Procedure{Poisoning}{$S_{tr}, S_{val}, \epsilon, \lambda, q$}
\State $t \gets 0$
\State $S_p^{(t)} = \{\textbf{x}_{p_j}^{(t)}, y_{p_j} \}_{j=1}^q \gets chooseInitialPoints(S_{tr})$
\Repeat
	\State ${\hat S}_p = \{ {\bf x}_{p_j}, y_{p_j} \}_{j=1}^q \gets S_p^{(t)}$
	\For{$j = 1,\dots,q$}
		\State $({\bf w}, b) \gets learnLasso(S_{tr} \cup {\hat S}_p)$
		\State $\text{Compute } \Delta({\bf x}_{p_j}) = \partial O_A/\partial {\bf x}_{p_j})$
		\State $\bm{g} = \Pi_{\X}({\bf x}_{p_j} + \Delta({\bf x}_{p_j})) - {\bf x}_{p_j}$
		\State $\eta \gets GS(S_{tr}, S_{val}, {\hat S}_p, \bm{g})$
		\State ${\bf x}_{p_j} \gets {\bf x}_{p_j} + \eta \ \bm{g}^T$
	\EndFor
	\State $t \gets t + 1$
	\State $S_p^{(t)} = \{\textbf{x}_{p_j}^{(t)}, y_{p_j} \}_{j=1}^q \gets {\hat S}_p$
\Until{$|O_A(S_p^{(t)}) - O_A(S_p^{(t-1)})| < \epsilon$}
\Return $S_p$
\EndProcedure
\end{algorithmic}
\caption{Poisoning Lasso-Like Linear Classifiers}
\label{alg1}
\end{algorithm}

\section{Detection of Poisoning Attacks with Outlier Detection}
To maximally influence the learning algorithm with a limited number of poisoning points, the attacker aims to achieve the maximum impact on the learner with each malicious point, as described by the bilevel optimization problem in (\ref{eq3}). In the case of linear classifiers, the loss function for the poisoning points can be made arbitrarily large by increasing $|{\bf w}^\top {\bf x}_{p_j}|$. Thus, if no restrictions are imposed on the possible values for the features of the poisoning points, it is expected that most of the points generated by the optimal attack strategy described in Sect.~3 are outliers.

\begin{figure}
\begin{center}
\includegraphics[width=0.7\columnwidth]{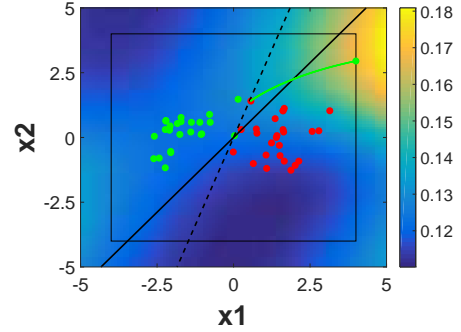}
\end{center}
\caption{Synthetic example: Poisoning a Lasso-like classifier. The colour map represents the MSE on the validation dataset. Dashed and solid black lines depict the decision boundary for the poisoned and clean dataset, respectively. The trajectory of the poisoning point during the gradient ascent strategy is shown in green. The black square determines the attacker's box constraint defined by $\mathcal{F} =  [-4, 4] \times [-4, 4]$.}
\label{fig1}
\end{figure}

To illustrate this point in Fig.~\ref{fig1} we depict a synthetic problem with 2 features where we aim to poison a Lasso-like classifier by injecting one malicious sample solving the bilevel optimization problem in (\ref{eq3}). In the example we have generated a binary classification problem where the distribution of negative and positive classes are bi-variate Gaussian distributions with covariance $\bm{\Sigma} = 0.6 {\bf I}$ and means $\bm{\mu}_{\pm} =\pm (1.5, 0)$ respectively. We have drawn $25$ samples for training and $100,000$ for validation from each of the two Gaussian distributions. For the purpose of the example we have constrained the samples to lie in the box defined by $\mathcal{F} =  [-4, 4] \times [-4, 4]$. Thus, the attacker uses the projector operator $\Pi_{\X}$ in (\ref{eq5}) to constrain and project the poisoning points onto the feasible domain determined by $\mathcal{F}$. The regularization parameter for the linear classifier has been set to $\lambda = 0.01$. In Fig.~\ref{fig1} we can observe that, by adding just one poisoning point to the training set, the decision boundary changes significantly. The color map in the background represents the cost on the validation dataset as a function of the position of the poisoning point. We can observe that the validation cost, which the attacker aims to maximize, increases as the poisoning point moves away from the genuine data distribution. Thus, the resulting poisoning point lies on the boundaries of the box constraint $\mathcal{F}$. It is also clear that if we do not define constraints for the attacker, the value of the features for the optimal poisoning point can be unbounded. Nevertheless, even defining such constraint, in the example in Fig.~\ref{fig1} we can observe that the optimal poisoning point, labelled as green, is still quite different from the genuine training points in the same (green) class. So although we can significantly affect the behavior of the learning algorithm just by adding a small fraction of poisoning points, if no reasonable constraints are imposed for the attacker, the generated attack points can be detected with appropriate pre-filtering, as we will show next. 

The poisoning strategies proposed in \cite{biggio12-icml,huang_icml15, mei2015using} do not consider any special restriction on the attacker's side to craft the poisoning points. Therefore, the generated adversarial examples only need to lie within a feasible domain, determined by the values that the different features can take, but not considering the underlying data distribution. The defined feasible space for the poisoning points does not consider the detectability constraints that can be expected from a sophisticated attacker. Outlier detection in high-dimensional spaces, i.e. when the number of features is large, is far from a trivial task, especially if the number of available samples is small compared to the number of features. In these settings, it is difficult to estimate the underlying probability distribution of the data. This is known as the \emph{curse of dimensionality} \cite{keogh2011curse}. However, although comparing distances in high-dimensional spaces is also challenging, we found that distance-based outlier detection techniques produce reliable indicators to detect adversarial training examples. 

\begin{figure}[t!]
\begin{center}
\includegraphics[width=1.0\columnwidth]{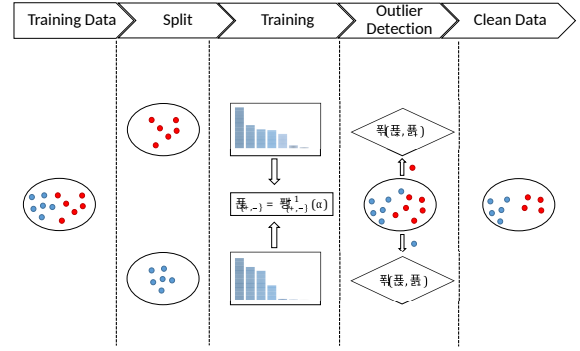}
\caption{Flowchart of the proposed defence algorithm. The training stage outputs a pair of thresholds to use at test time to identify outliers.} 
\label{fig4:1}
\end{center}
\end{figure}

In many practical applications, machine learning systems are re-trained from time to time to adapt to changes in the distribution of the data as new samples are collected. Given the huge amount of data collected by many applications, the curation of the training data is not always possible and poisoning attacks represent a severe threat. However, curation of a small fraction of the data can be achievable. Under these settings, we propose to use distance-based anomaly detection to detect adversarial training examples using a small fraction of \emph{trusted} data points. In Fig.~\ref{fig4:1} we depict the flowchart of the proposed defence against the poisoning attacks in \cite{biggio12-icml,huang_icml15, mei2015using}. We first split a small fraction of trusted data $D$ into the different classes, i.e. $D_+$ and $D_-$. Then, we train one distance-based outlier detector for each class using this curated data. Given a dataset, most outlier detection algorithms provide an \emph{outlierness score}, $q(\bf{x})$ for each $\bf{x}$ in the dataset. To compute the threshold to detect outliers based on $q(\bf{x})$ we use the \emph{Empirical Cumulative Distribution Function} (ECDF), $\hat{F}$, of the training instances' scores. Then, we identify the $\alpha$-percentile, i.e. the value $t$ such that $\hat{F}^{-1}(\alpha) = t$. The $\alpha$-percentile controls the fraction of genuine points that are expected to be retained after applying the outlier detector. Smaller values of $\alpha$ will reduce the number of genuine points to be considered for training, which can affect the accuracy of the classifier. However, a small $\alpha$ decreases the accuracy to detect poisoning points. Once the thresholds, $t_-$ and $t_+$, are calculated for the positive and negative classes\footnote{Note that a similar scheme can also be applied for multi-class classification problems.} and a new (untrusted) dataset $D'$ is collected for re-training the learning algorithm, we remove all the samples ${\bf x}' \in D': q(\bf{x}') > t_{c({\bf x})}$; where $c({\bf x})$ is the label of $\bf{x}$. It is important to note that this defence mechanism works if the data used to train the outlier detector is trusted. If this assumption is violated, the outlier detector can be poisoned as well. In those cases, defending against adversarial examples is even harder since the adversary can be free to significantly change the underlying data distribution \cite{Kearns93learningin}. 

\begin{table*}[t!]
\begin{center}
\begin{tabular}{|c|c|c|c|c|}
\hline
\multicolumn{2}{|c|}{\textbf{Score Function}} & \textbf{Parameters} & \textbf{Complexity} & \textbf{Reference} \\
\hline
$q_{k}(\bf{x})$ & $d^k({\bf x}; D)$ & $k = 5$ & $O(nmd)$ & \cite{ramaswamy2000efficient} \\
$q_{kSp}(\bf{x})$ & $d^k({\bf x}; S_{\bf{x}}(D))$ & $k=5, s= 20$ & $ O(mds)$  & \cite{wu2006outlier}  \\
$q_{Sp}(\bf{x})$ & $\min_{\textbf{x}'\in S(D)} d({\bf x}, {\bf x}')$ & $s= 20$ & $ O(mds)$ &  \cite{sugiyama2013rapid} \\
$q_{SVM}(\bf{x})$ & $\rho - {\bf x}^\top\bf{w}$ & $C = CV$ & $ O(md)$ & \cite{scholkopf2001estimating} \\
$q_{LOF}({\bf x})$ & $(|N^k({\bf x})|^{-1}\sum_{{\bf x}' \in N^k({\bf x})} \rho({\bf x}'))\rho({\bf x})^{-1}$ & $k=5$ & $ O(n^2md)$ & \cite{breunig2000lof} \\
\hline
\end{tabular}
\end{center}
\caption{Summary of the considered outlier detection algorithms. The parameters are chosen accordingly to the authors suggestions. In the case of $q_{SVM}({\bf x})$, we choose the $C$ based on leave-one-out cross validation. The complexity entry refers to the time needed to compute the score for each point in the set $D$.}
\label{tab4:1}
\end{table*}

For outlier detection, we have compared a number of popular algorithms based on the distance between data points \cite{ramaswamy2000efficient,wu2006outlier,sugiyama2013rapid,scholkopf2001estimating} and a well-known distribution-estimation based method \cite{breunig2000lof}. The score functions of these methods and their computational complexity are reported in \Cref{tab4:1} along with the values of their parameters used in the experimental evaluation in Sect.~5 (using the values recommended in the corresponding papers). Given a dataset $D$ with $n$ data points $\{{\bf x}_i \}_{i = 1}^n$ with ${\bf x}_i \in \R^d$, the outlier detector in \cite{ramaswamy2000efficient} uses the euclidean distance of a test sample $\bf{x}$ to the $k$-th nearest neighbour in $D$ (denoted as $d^k({\bf x}; D)$ in \Cref{tab4:1}) as outlierness score. Similarly, \cite{wu2006outlier} computes the euclidean distance of $\bf{x}$ to the $k$-th nearest neighbour from a subset of $s$ points $S_{{\bf x}}(D)$ sampled without replacement from $D$. In contrast, in \cite{sugiyama2013rapid} the outlierness score is given by the euclidean distance between the test point and the nearest neighbour given a subset $S(D)$ with $s$ data points sampled with replacement from $D$. The algorithms in \cite{wu2006outlier,sugiyama2013rapid} are computationally  more efficient than \cite{ramaswamy2000efficient}, as they only require to compute the distance w.r.t. a reduced number of data points, making them more appropriate for large datasets. The one-class SVM approach described in \cite{scholkopf2001estimating} aims to find the maximum margin hyperplane that separates the training data from the origin in the feature space determined by a specified kernel (in our experimental evaluation we have not applied any kernel). This maximum margin hyperplane corresponds to a decision boundary that encloses most of the training data. The outlierness score is computed as $\rho - \bf{x}^\top\bf{w}$, where $\rho$ and $\bf{w}$ are the solution of the quadratic optimization problem of the corresponding soft-margin SVM formulation in \cite{scholkopf2001estimating}. Finally, the method proposed in \cite{breunig2000lof} is based on the estimation of the \emph{local reachability density} $\rho(\bf{x})$. This quantity grows when $\bf{x}$ and its neighbourhood are close to the other points of $D$. The outlierness score is computed as the ratio between the average density of $N^k(\bf{x})$ and $\rho(\bf{x})$. Then, if a point is far from their neighbours, even if these neighbours are close to other points in $D$, the outlierness score is large.

\section{Experimental Evaluation}
We have performed an experimental evaluation to assess the validity of our defence using outlier detection  to mitigate the optimal attack strategy, described in Section~3, against linear classifiers. For this, we have selected two real datasets from the UCI repository:\footnote{\url{https://archive.ics.uci.edu/ml/datasets.html}} \,\emph{Spambase} and \emph{MNIST}, which are common benchmarks in spam filtering and computer vision, which are typical application domains in the literature on adversarial machine learning. In the experimental evaluation we have also included label flipping attack strategies, which have also been shown to be effective to compromise machine learning classifiers, although they are usually less effective than the optimal attack strategies \cite{biggio12-icml}. We have tested our defence using the different outlier detection algorithms described in Sect.~4, exploring their sensitivity w.r.t. the percentile of the ECDF threshold.

\begin{table*}[t!]
\begin{center}
\subfloat[SpamBase]{\begin{tabular}{|c|c|ccccc|}
\hline
\textbf{Method} & $\alpha$\textbf{-percentile} & \multicolumn{5}{|c|}{\textbf{\% of Poisoning Points}} \\
 & & 0\% & 5\% & 10\% & 15\% & 20\% \\
\hline
No defence & - &$0.112 \pm 0.010$ & $0.16 \pm 0.015$ &  $0.174 \pm 0.019$ & $0.185 \pm 0.020$ & $0.195 \pm 0.019$ \\
\hline
\multirow{3}{*}{$q_{k}$}       & $90$ & $0.117 \pm 0.013$ & $0.117 \pm 0.013$ & $0.117 \pm 0.013$ & $0.117 \pm 0.013$ & $0.117 \pm 0.013$\\
 				           & $95$ & $0.114 \pm 0.012$ & $0.114 \pm 0.012$ & $0.114 \pm 0.012$ & $0.114 \pm 0.012$ & $0.114 \pm 0.012$\\
 				           & $99$ & $0.114 \pm 0.012$ & $0.114 \pm 0.012$ & $0.114 \pm 0.012$ & $0.114 \pm 0.012$ & $0.114 \pm 0.012$\\
\hline
$q_{kSp}$ 		           & $99$ & $0.113 \pm 0.010$ & $0.113 \pm 0.010$ & $0.114 \pm 0.012$ & $0.117 \pm 0.010$ & $0.120 \pm 0.012$\\
\hline
$q_{Sp}$  			           & $99$ & $0.113 \pm 0.010$ & $0.113 \pm 0.010$ & $0.114 \pm 0.010$ & $0.115 \pm 0.009$ & $0.116 \pm 0.011$\\
\hline
\multirow{3}{*}{$q_{SVM}$} & $90$ & $0.113 \pm 0.010$ & $0.124 \pm 0.010$   & $0.126 \pm 0.011$ & $0.134 \pm 0.0140$ & $0.137 \pm 0.015$\\
 				           & $95$ & $0.115 \pm 0.010$ & $0.123 \pm 0.011$ & $0.125 \pm 0.013$ & $0.131 \pm 0.0110$ & $0.136 \pm 0.015$\\
 				           & $99$ & $0.114 \pm 0.010$ & $0.122 \pm 0.010$   & $0.126 \pm 0.013$ & $0.127 \pm 0.0110$ & $0.137 \pm 0.015$\\
\hline
\multirow{3}{*}{$q_{LOF}$}  & $90$ & $0.119 \pm 0.007$ & $0.119 \pm 0.007$ & $0.119 \pm 0.007$ & $0.119 \pm 0.007$ & $0.119 \pm 0.007$\\
 				           & $95$ & $0.115 \pm 0.007$ & $0.115 \pm 0.007$ & $0.115 \pm 0.007$ & $0.115 \pm 0.007$ & $0.115 \pm 0.007$\\
 				           & $99$ &\bm{$0.111 \pm 0.009$} & \bm{$0.111 \pm 0.009$} & \bm{$0.111 \pm 0.009$} & \bm{$0.111 \pm 0.009$} & \bm{$0.111 \pm 0.009$}\\
\hline
\end{tabular}} \\
\subfloat[MNIST]{\begin{tabular}{|c|c|ccccc|}
\hline
\textbf{Method} & $\alpha$\textbf{-percentile} & \multicolumn{5}{|c|}{\textbf{\% of Poisoning Points}} \\
 & & 0\% & 5\% & 10\% & 15\% & 20\% \\
\hline
No defence & - & $0.037 \pm 0.005$ & $0.087 \pm 0.034$ & $0.210 \pm 0.146$ & $0.296 \pm 0.164$ & $0.391 \pm 0.160$ \\
\hline
\multirow{3}{*}{$q_{k}$}        & $90$ & $0.044 \pm 0.004$ & $0.066 \pm 0.017$ &  $0.074 \pm 0.021$ &  $0.073 \pm 0.017$ &  $0.079 \pm 0.025$  \\
 				           & $95$ & $0.042 \pm 0.004$ & $\bm{0.061 \pm 0.012}$ &  $\bm{0.069 \pm 0.017}$ &  $0.069 \pm 0.014$ &  $\bm{0.07   \pm 0.012}$  \\
 				           & $99$ & $0.039 \pm 0.006$ & $0.062 \pm 0.013$ &  $0.070 \pm 0.014$   &  $0.069 \pm 0.013$ &  $\bm{0.070 \pm 0.013}$  \\
\hline
$q_{kSp}$                            & $99$ & $\bm{0.038 \pm 0.004}$ & $\bm{0.061 \pm 0.012}$ & $0.076 \pm 0.029$ & $0.067 \pm 0.011$ & $0.071 \pm 0.012$ \\
\hline
$q_{Sp}$                              & $99$ & $0.039 \pm 0.005$ & $0.062 \pm 0.013$ & $0.072 \pm 0.019$ & $\bm{0.068 \pm 0.011}$ & $0.070 \pm 0.014$ \\
\hline
\multirow{3}{*}{$q_{SVM}$}  & $90$ & $0.039 \pm 0.004$ & $0.062 \pm 0.016$ &  $0.072 \pm 0.002$ &  $0.073 \pm 0.018$ &  $0.079 \pm 0.002$  \\
 				            & $95$ & $0.039 \pm 0.005$ & $0.061 \pm 0.016$ &  $0.073 \pm 0.002$ &  $0.073 \pm 0.019$ &  $0.079 \pm 0.019$  \\
 				            & $99$ & $0.039 \pm 0.005$ & $0.061 \pm 0.016$ &  $0.072 \pm 0.002$   &  $0.07 \pm 0.014$ &  $0.078 \pm 0.019$  \\
\hline
\multirow{3}{*}{$q_{LOF}$}   & $90$ & $0.049 \pm 0.007$ & $0.088 \pm 0.035$ &  $ 0.108 \pm 0.037$ &  $0.146 \pm 0.128$ &  $0.155 \pm 0.123$  \\
 				            & $95$ & $0.044 \pm 0.005$ & $0.074 \pm 0.002$ &  $ 0.086 \pm 0.023$ &  $0.087 \pm 0.002$ &  $0.100 \pm 0.039$  \\
 				            & $99$ & $0.041 \pm 0.006$ & $0.066 \pm 0.017$ &  $ 0.075 \pm 0.016$ &  $0.073 \pm 0.001$ &  $0.079 \pm 0.022$  \\
\hline
\end{tabular}}
\end{center}
\caption{Results on SpamBase and MNIST datasets. Averaged test classification error plus/minus the standard deviation as a function of the poisoning points. The first row shows the effect of poisoning when no countermeasure is applied. For each outlier detection method we show the results for three different $\alpha$-percentiles: $90$, $95$ and $99$. For $q_{kSp}, q_{Sp}$ the errors for the different thresholds are not statistically different, as a result of a paired $t$-test. Then, we only report the results for the $99$-th percentile. In bold, the results of the best method in the dataset for a given poisoning level.}
\label{tab5:1}
\end{table*}

\subsection{Spambase Dataset}
\emph{Spamabase} is a dataset for spam filtering benchmarking consisting of $4,601$ emails out of which $1,813$ are spam (the positive class) and $2,788$ ham, i.e. good emails (the negative class). Emails are represented by means of 57 features: 3 of them representing word statistics and the remaining representing term frequencies of the most significant words. In our experiments we have restricted the feature set considering only the term frequencies (54 features). On this reduced feature set the baseline accuracy, i.e., the one obtained in the absence of attack, is not affected, but the poisoning strategy is easier to apply. We have converted the frequencies into a bag-of-word model (i.e. we have used binary features ${\bf x}'$ such that $x_i' = 1$ if $x_i > 0$, otherwise $x_i' = 0$) the so the resulting problem is better separated. Moreover, we have removed all duplicated examples. Our final dataset contains $2,443$ and $1,657$ good and spam emails respectively. For the experiments we have created 10 random splits of the data with 200 examples to train the classifier, 200 examples to train the outlier detection, 400 examples for the attacker's validation set and the remaining data for testing. In the case of the random outlier detection methods, $q_{kSp}$ and $q_{Sp}$, for each split we have averaged the results over 10 repetitions. In this case, since the number of features is relatively small compared to the training set size, we do not regularize the linear classifier (i.e. we set $\lambda = 0$). Finally, to compute the attack points, we treat the features as continuous in order to compute the gradients in the bi-level optimization problem. Then, we round them at the end of the optimization procedure, similarly to \cite{biggio12-icml,Munoz-Gonzalez:2017:TPD:3128572.3140451}.

The results are reported in \Cref{tab5:1}.(a), showing the effect of the poisoning attack as a function of the fraction of poisoning points (e.g. 5\% poisoning for a training set of 200 examples corresponds to adding 10 malicious examples). We can observe that when no defence is applied the classification error on the test set increases from $0.112$ to $0.195$ with 20\% of poisoning points. On the other hand, it can be appreciated that, regardless of the specific method, outlier detection is very effective at mitigating the effect of the attack. Actually, the distribution-based approach, $q_{LOF}$, even improves the performance on the clean dataset when we set the outlier detection threshold to the 99 percentile. The distance-based algorithms achieve similar performance, but at a reduced computational cost. It is interesting to observe that for both $q_{kSp}$ and $q_{Sp}$ the performance is insensitive to the specific threshold as confirmed by a paired $t$-test analysis at a $0.99$ confidence level. Thus, even if some non-malicious samples are removed, the performance of the classifier is not affected.


\subsection{MNIST Dataset}
\emph{MNIST} is a dataset for handwritten digit recognition \cite{mnistRef}. Although the complete dataset involves 10 different digits, as in \cite{biggio12-icml}, we consider the subtask of distinguishing between digits 7 and 1 which is a binary classification problem. Each digit is represented as a feature vector representing a $28 \times 28$ gray-scaled image, where each one of the 784 features represents the relative pixel intensity on a scale between $[0, 255]$. We normalize each feature to be in the range $[0, 1]$. The experimental settings are similar to the case of \emph{Spambase}, but in this case we use 800 samples for training the classifier, 800 for training the outlier detector, $1,000$ for the attacker's validation set, and the remaining for testing. In this case we select the regularization parameter $\lambda$ by means of a 5-fold cross validation procedure on the training data poisoned with the initial value of the poisoning points (i.e. the \emph{warm start}). 
When $\lambda$ is chosen with a data-dependent procedure, as cross validation, its value is also affected by the addition of the poisoning points. Ideally, the poisoning attack strategy should consider this dependency, but none of the previous work in the literature has modelled this aspect. Then, for our experimental comparison with the optimal attack strategy in \cite{huang_icml15} we only provide a value for $\lambda$ regardless of the fraction of poisoning points. We leave the investigation of the evolution of $\lambda$ as a function of the poisoning points for future research.

The experimental results on MNIST dataset are reported in \Cref{tab5:1}.(b). In this case, due to the relatively small training set, compared to the number of features, the poisoning attack is even more effective. The error raises from $0.037$ in the clean dataset to $0.391$ after adding 20\% of poisoning points, i.e. the error increases more than 10 times. In this case, the standard deviation of the errors is quite large when no defence is applied. This is possibly due to the sub-optimality of the attack in \cite{huang_icml15} which assumes that $\lambda$ is constant regardless of the fraction of poisoning points. As in the previous case, the results in \Cref{tab5:1}.(b) show the effectiveness of our defence with outlier detection to mitigate the effect of the poisoning attacks. For example in the case of 20\% poisoning the best method, $q_{Sp}$, leads to a classification error of about $0.07$, whereas the error rate when no defence is applied is $0.391$. It is also interesting to note that $q_{LOF}$, which performed best in the previous experiment, now offers the worst performance amongst the different outlier detection methods compared. We think that this is due to the difficulty of doing density estimation accurately in high-dimensional spaces. 

\begin{figure*}[t!]
\begin{center}
\psfrag{Opt}{{\scriptsize Opt}}
\psfrag{OD}{{\scriptsize OD}}
\psfrag{ILF}{{\scriptsize ILF}}
\psfrag{RLS}{{\scriptsize RLS}}
\psfrag{RLF}{{\scriptsize RLF}}
\psfrag{Classification Error}{{\scriptsize Classification Error}}
\psfrag{\% poisoning samples}{{\scriptsize \% poisoning samples}}
\subfloat[][Optimal attack (Spambase)]{\includegraphics[width=5cm, height=4.0cm]{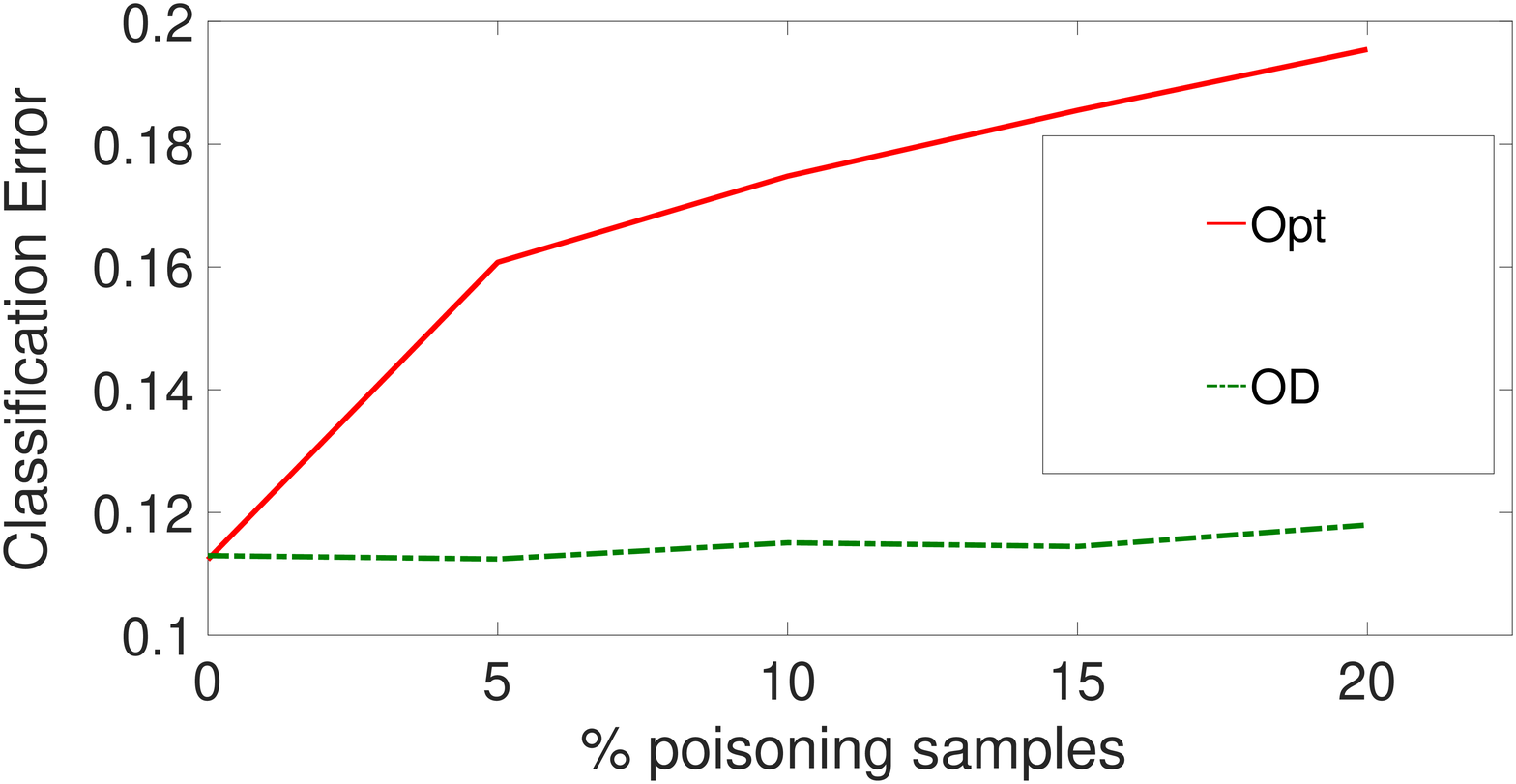}}
\quad
\subfloat[][ILF (Spambase)]{\includegraphics[width=5cm, height=4.0cm]{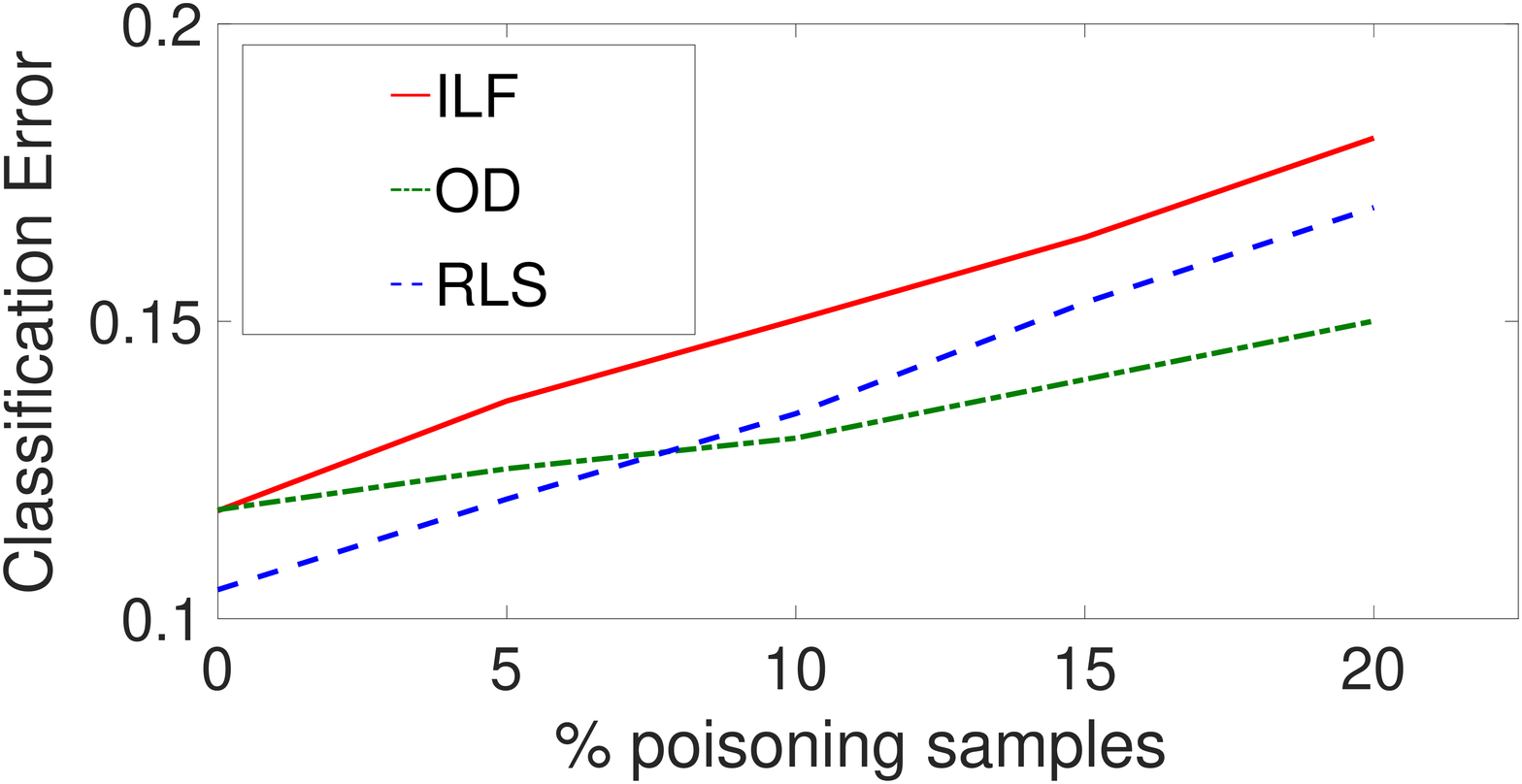}}
\quad
\subfloat[][RLF (Spambase)]{\includegraphics[width=5cm, height=4.0cm]{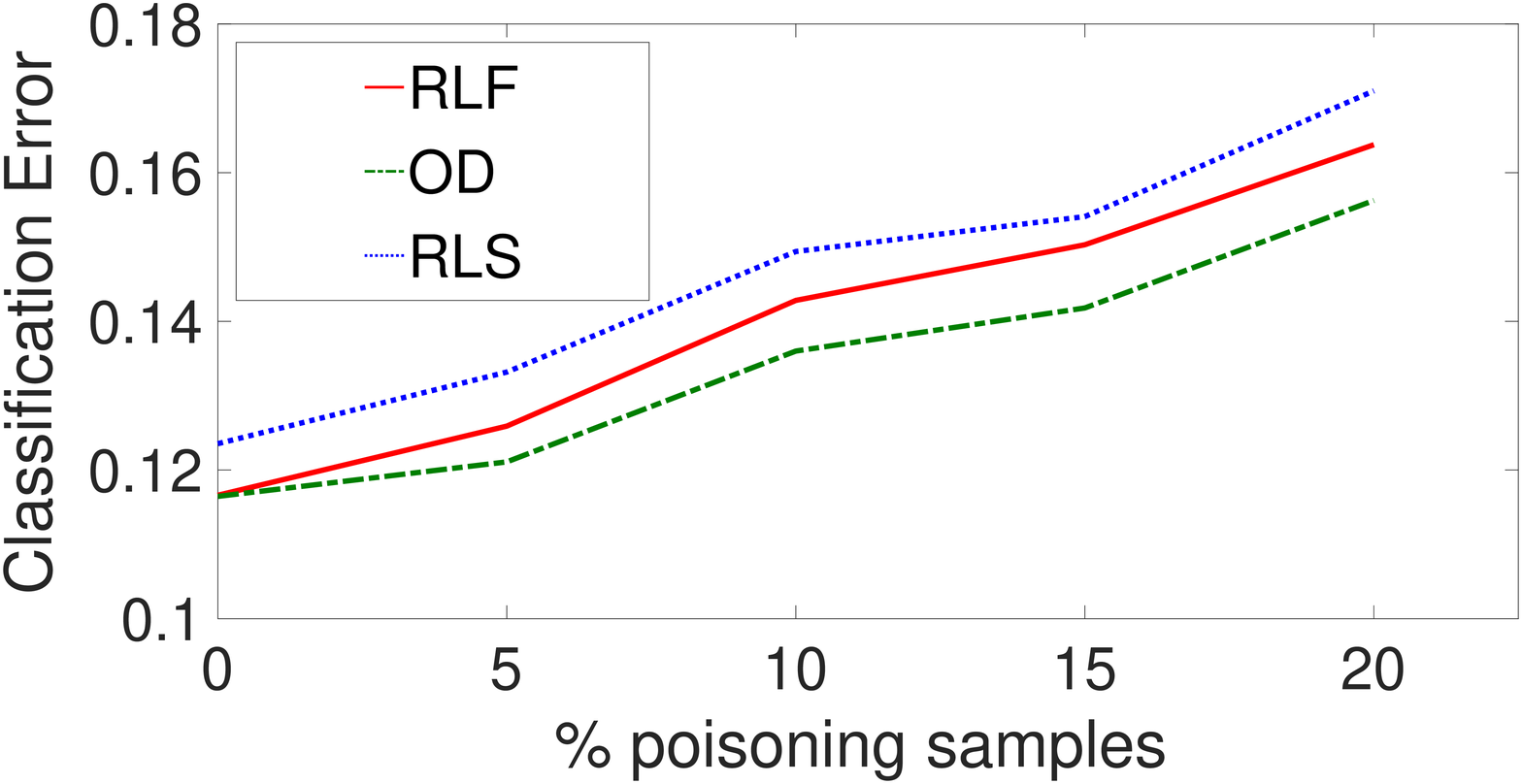}}
\\
\subfloat[][Optimal attack (MNIST)]{\includegraphics[width=5cm, height=4.0cm]{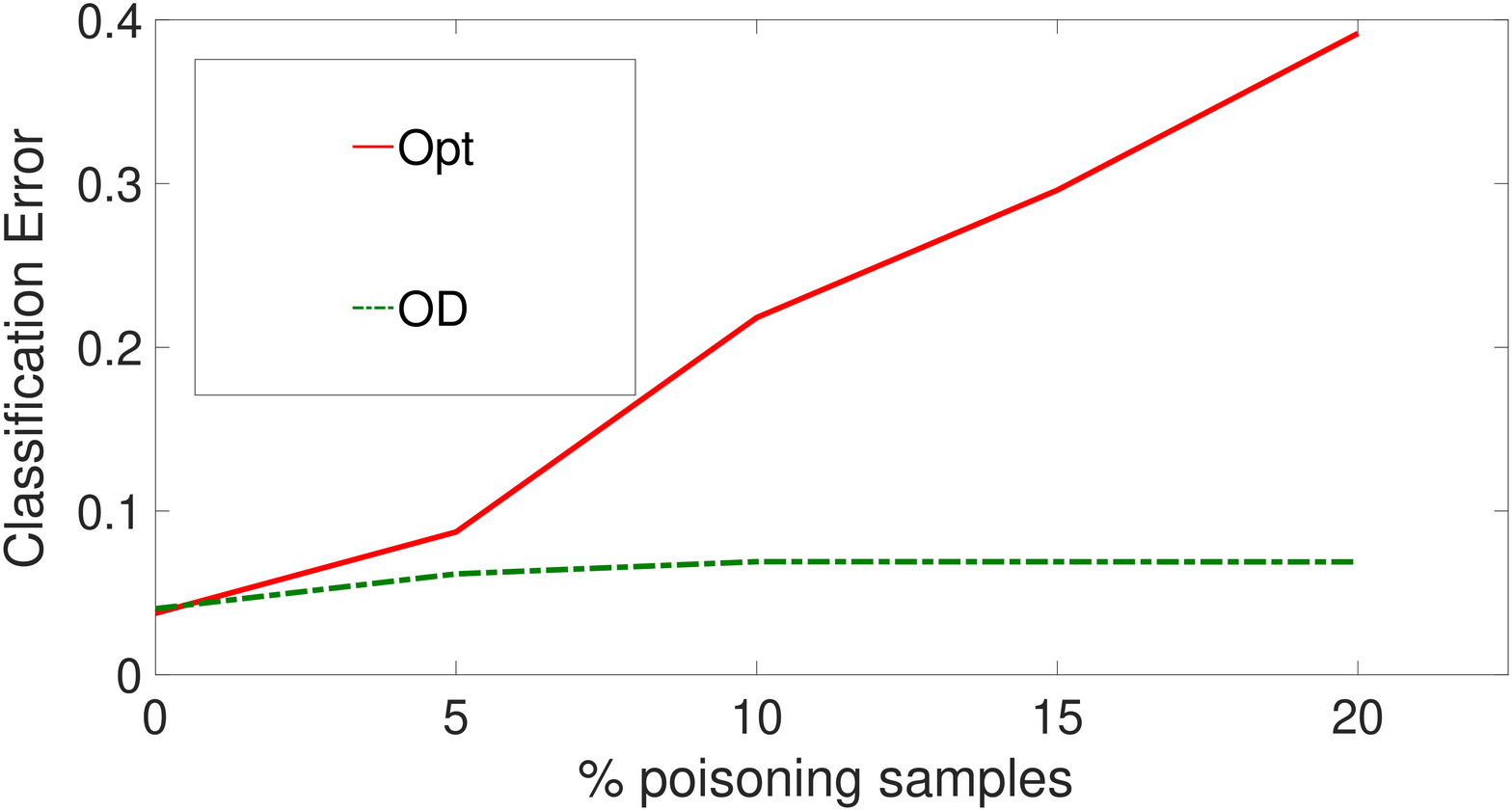}} 
\quad
\subfloat[][ILF (MNIST)]{\includegraphics[width=5cm, height=4.0cm]{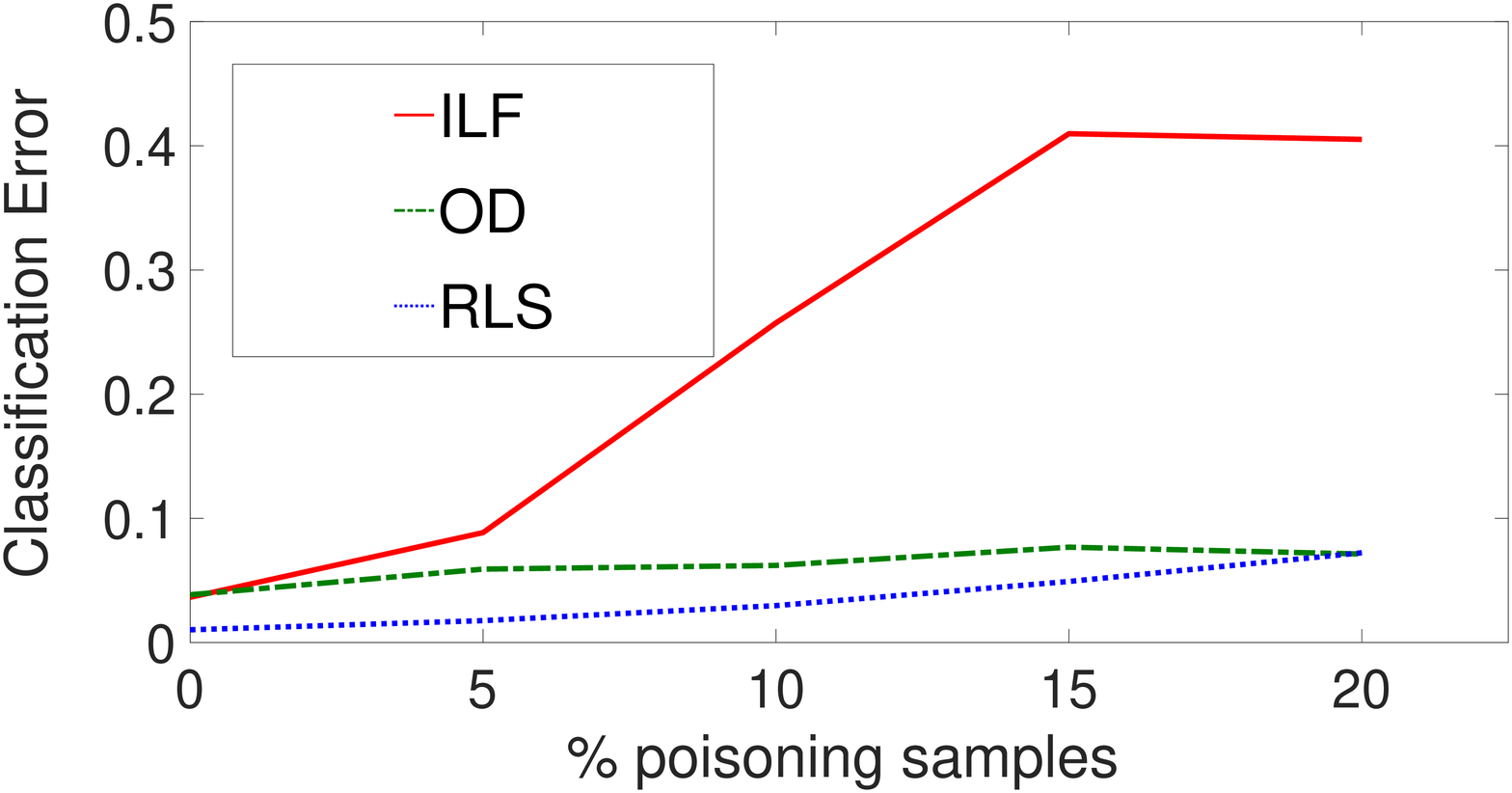}}
\quad
\subfloat[][RLF (MNIST)]{\includegraphics[width=5cm, height=4.0cm]{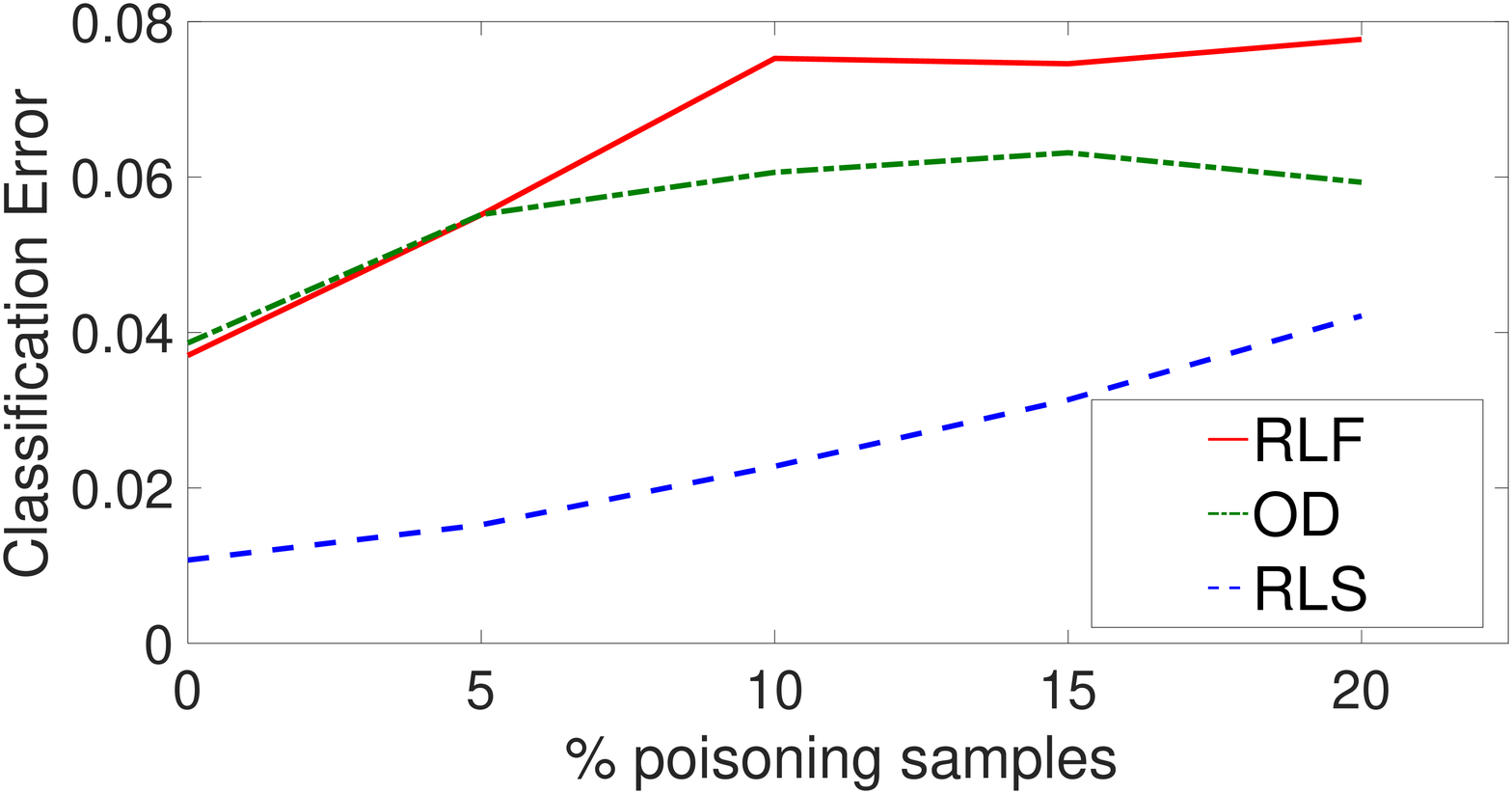}} 
\end{center}
\caption{Evaluation of outlier detection and RLS defences against optimal poisoning attacks and label flipping attacks. Each plot reports the average test classification error as a function of the percentage of poisoning points. The solid red line shows the classification error when no defence is applied. Dashed-dotted green lines depict the classification error for our outlier detection based countermeasure (OD in the legend) when using as threshold of $0.99$. The dashed blue lines depict the classification error of RLS. \emph{First Row:} Spambase. \emph{Second Row:} MNIST.}
\label{fig5:1}
\end{figure*}

\subsection{Label Flipping}
We have also evaluated the effect of \emph{label flipping} strategies. Compared to the optimal attack, label flipping is more constrained, as the adversary can only change the label of the samples. We have considered two variants of label flipping attacks, namely \emph{Random} (RLF) and \emph{Informed} (ILF) \emph{Label Flipping}. In RLF, the attacker selects points at random from the training dataset and flips their labels. In contrast, in ILF the attacker selects the training examples showing the largest squared error with respect to a classifier learned over the clean training data when their label is flipped. This is similar to the approach used in \cite{biggio12-icml} to select the initial points for optimal poisoning attacks. In our experiment we have only considered $q_{Sp}$ as outlier detection method with a threshold equal to the $99$-th percentile of the empirical CDF, as it showed good performance to mitigate the optimal poisoning attack on both datasets, as reported in \Cref{tab5:1}, and it seems quite insensitive to the threshold. We have used the same 10 random splits generated for the previous experiment and, for each split, we have performed 10 repetitions. In the case of RLF, for each split we also perform 10 further repetitions in order to cope with the randomness of the selection of the examples to flip. Thus, each combination of parameters requires $1,000$ repetitions in total. In this experiment, we have also compared our countermeasure against the method of \emph{unbiased estimators} proposed by \cite{natarajan2013learning}. We refer to this method as RLS. This method is specifically tailored to mitigate label flipping attacks. It learns a classifier by minimizing the empirical risk derived from the robust loss $\tilde{\ell}(h, ({\bf x}, y)) = ((1-\rho_{-y})\ell(h, ({\bf x}, y)) - \rho_y\ell(h, ({\bf x}, -y)))/(1-\rho_{+1} - \rho_{-1})$. We have considered $\ell$ to be the squared loss and we have used gradient descent to optimize the empirical loss. We have selected a learning rate of $0.1$ and $0.01$ for Spambase and MNIST respectively. This algorithm requires to specify two noise levels, $\rho_+$ and $\rho_-$, for the positive and the negative examples, which are typically unknown. As suggested in \cite{natarajan2013learning} we have estimated these noise levels with a 5-fold cross validation procedure exploring the values given in the set $\{0, 0.05, 0.1, 0.15, 0.2\}$. For RLF, we have selected $\rho = \rho_+ = \rho_-$.

\Cref{fig5:1} shows the average test classification error as a function of the poisoning points for the optimal and label flipping attacks. For the sake of clarity we have omitted the error bars in the plots. 
Similar to our previous results, in \Cref{fig5:1}.(a) and (d) we can observe that outlier detection is effective to mitigate the optimal attack in both datasets. In the case of ILF in \Cref{fig5:1}.(b) and (e) both outlier detection and RLS are capable of mitigating the attack in MNIST dataset. Actually, RLS is capable of outperforming the standard classifier when evaluated on the clean dataset. This can happen because RLS removes samples that, even if they are genuine, can be considered harmful for the classifier. However, the effect of both defence mechanisms to mitigate the poisoning attack in Spambase dataset is limited, although the degradation of the performance as a function of the poisoning points is more graceful for the outlier detection defence. We can also observe that, as RLF is a less aggressive attack strategy and, hence, the effect of the poisoning attack is reduced compared to ILF and the optimal attack, the malicious points are more difficult to detect for both outlier detection and RLS. Thus, on average, the attack points are closer to the genuine points of the same class. For example, in Spambase, with 20\% of poisoning points, the averaged test error after applying outlier detection is $0.149$, whereas when no defence is applied the error is $0.180$. In \Cref{fig5:1}.(c) and (f) we can also observe that the RLS defence partially mitigates the effect of the attack for MNIST, whereas in Spambase the average performance is even worse than the poisoning attack itself. The experimental results show that label flipping strategies produce attack points that are less trivial, compared to the optimal attack strategies. Thus, it is more difficult to detect them, although the effect of the attack is less significant when no defence is applied. This suggests that there exists a trade-off between effectiveness and detectability of the attack that should be considered when modeling poisoning attack strategies.

\section{Conclusion}
Poisoning attacks are considered one of the most relevant emerging threats against machine learning and data-driven technologies. Since many applications rely on untrusted data collected in the wild, attackers can inject malicious data capable of degrading the performance of the system in a targeted or indiscriminate way. Optimal attack strategies have been proposed against popular machine learning classifiers \cite{biggio12-icml,huang_icml15, mei2015using,Munoz-Gonzalez:2017:TPD:3128572.3140451}. Although they have been shown to be very effective, the proposed algorithms do not include any explicit mechanism to model detectability constraints. The attack points generated by these strategies are far from the genuine points, and then, can be easily removed with adequate pre-filtering of the data. In this paper we have proposed an outlier-detection-based scheme capable to detect the attack points against linear classifiers. Although performing outlier detection is challenging in high-dimensional datasets, we have shown empirically that these techniques are capable of mitigating the effect of optimal poisoning attack strategies. At the same time, our experimental evaluation shows that less aggressive attacks, such as label flipping, can be difficult to detect with these defence mechanisms, since the attack points generated are, on average, closer to the genuine data points. This suggests the importance of modeling detectability constraints when designing poisoning attack strategies, similar to other applications in cyber security, where there is a natural trade-off between the damage of the attack and its detectability. Future work includes exploration and adoption of aggressive localization, unlabelled data and margin optimization techniques, which have been successfully applied in computational learning theory for similar problems.

\bibliography{mybibfile}

%
%
%

\end{document}